# A personal model of trumpery: Deception detection in a real-world high-stakes setting

**Authors:** Sophie Van Der Zee[1†], Ronald Poppe[2], Alice Havrileck[1 3], Aurélien Baillon[1].

**Affiliations:**

[1] Department of Applied Economics, Erasmus School of Economics, Erasmus University Rotterdam, Rotterdam, The Netherlands;

[2] Information and Computing Sciences, Utrecht University, Utrecht, The Netherlands;

[3] Department of Economics and Management, École Normale Supérieure Paris-Saclay, Cachan, France

† Corresponding address:

vanderzee@ese.eur.nl

Erasmus University Rotterdam, Burgemeester Oudlaan 50, 3062 PA Rotterdam, The Netherlands

**Language use reveals information about who we are and how we feel[1-3]. One of the pioneers in text analysis, Walter Weintraub, manually counted which types of words people used in medical interviews and showed that the frequency of first-person singular pronouns (i.e., I, me, my) was a reliable indicator of depression, with depressed people using *I* more often than people who are not depressed[4]. Several studies have demonstrated that language use also differs between truthful and deceptive statements[5-7], but not all differences are consistent across people and contexts, making prediction difficult[8]. Here we show how well linguistic deception detection performs at the individual level by developing a model tailored to a single individual: the current US president. Using tweets fact-checked by an independent third party (Washington Post), we found substantial linguistic differences between factually correct and incorrect tweets and developed a quantitative model based on these differences. Next, we predicted whether out-of-sample tweets were either factually correct or incorrect and achieved a 73% overall accuracy. Our results demonstrate the power of linguistic analysis in real-world deception research when applied at the individual level and provide evidence that factually incorrect tweets are not random mistakes of the sender.**

Since the early work by Weintraub[4], methods have been developed to automatically count the different types of words people use[9,10], making text analysis an efficient and objective research method to study stable traits such as personality[1-4] and more temporary states such as cooperation[11]. Thanks to substantial improvements in automated speech recognition, the usability and applicability of such analyses will only further increase[12]. Applied to deception detection, linguistic analysis showed that liars use more tentative, angry, emotional, and cognitive-processing words than truth tellers[13]. However, the observed patterns have been partly contradictory and have limited discriminative power[8]. One possible explanation is that the differences between truthful and deceptive language are too small to be consistently observed. Alternatively, there might be significant variation between individuals in language use, which limits the performance of one-size-fits-all models[14]. The question remains how well the techniques for a whole population can be tailored to a single person. Answering this question requires a large set of statements by a single individual of which the ground truth is known[15]. To date, this has proven challenging because the large scale fact-checking needed to acquire such a dataset requires substantial time and effort.

In the present paper, we put language-based lie detection to the test in a unique real-world setting, by analysing a multitude of statements made by a single individual, the current US president. Several organisations have tasked themselves with fact-checking statements of US presidents[16]. The communication channels that are checked include official speeches, Facebook posts, and tweets. Although some are posted under the president's name (e.g., Facebook posts), or are delivered by the president (e.g., speeches), the message itself may have been crafted by others, potentially introducing noise to the signal. Of all fact-checked communication channels, the current US president seems to be most in control of his Twitter account[17]. In addition, tweets are considered official White House communication[18]. This official status combined with systematic fact-checking provides the opportunity for deception detection for a single individual, in a high-stakes official context.

Previous analyses by independent fact-checkers concluded that presidential candidates and US presidents regularly say or write things that are factually incorrect[19,20]. In the media, these incorrect statements are often portrayed as deceptive[21]. However, an incorrect statement does not necessarily imply a lie. The sender may simply be wrong and have a false belief[22]. Being wrong should not affect language use because there is no difference in the perception or intention of the sender. In contrast, when deliberately presenting false statements as truths, one would expect a change in language use, according to the deception hypothesis[5-7,13] . Lying can cause behavioral change because it is cognitively demanding, elicits emotions, and increases attempted behavioral control[8,23]. We tested the deception hypothesis by investigating whether differences in language can be used to distinguish between correct and incorrect statements.

We collected a dataset (Dataset 1) of 605 presidential tweets by @realDonaldTrump over February – April 2018. The Washington Post provided us with a dataset comprising fact-checked communications from this period that we matched to our Twitter data file. As a result, we could identify which tweets were deemed incorrect, and labelled the remainder as correct. We named the obtained variable veracity. We screened the dataset to remove tweets contaminated by others (e.g., retweets, see details in the Methods section), resulting in a dataset containing 447 tweets (29.53% incorrect). Subsequently, we used the program Linguistic Inquiry and Word Count (LIWC2015) to identify how often different word categories were used in each tweet. LIWC calculates a score for 93 different categories, ranging from linguistic dimensions such as personal pronouns and negations, to psychological processes such as positive and negative emotions and cognitive processing[9]. During the screening process, we removed web links from the tweets, thereby altering punctuation in tweets. As a result, we excluded the LIWC punctuation categories except exclamation marks, leaving 82 categories. In addition, two dummy variables were created to identify whether a tweet included @ and # characters, since these are specifically relevant to tweets.

The deception hypothesis predicts that the type of words differs between truthful and deceptive statements. We ran a MANOVA comparing the means of the 82 LIWC categories and two dummy variables between correct and incorrect tweets. Using Pillai's trace, it revealed a main effect of veracity $V = .35$, $F\ (84, 362) = 2.37$, $p < 0.001$ (two-sided, as all the tests reported in this paper), $\eta^2_p = .36$ and demonstrated that 28 (33.33%) word categories significantly differed ($\alpha = 1\%$) between the correct and incorrect statements. Results are reported in detail in the Supplementary Information file and Figure 1 gives an overview of variables. Overall, correct statements contained more positive feelings (i.e., emotional tone and positive emotions) whereas incorrect statements were more evasive, with an increase in negations, tentative words, and comparisons. The sender used fewer # and @ in incorrect tweets, which reduces active engagement with the audience.

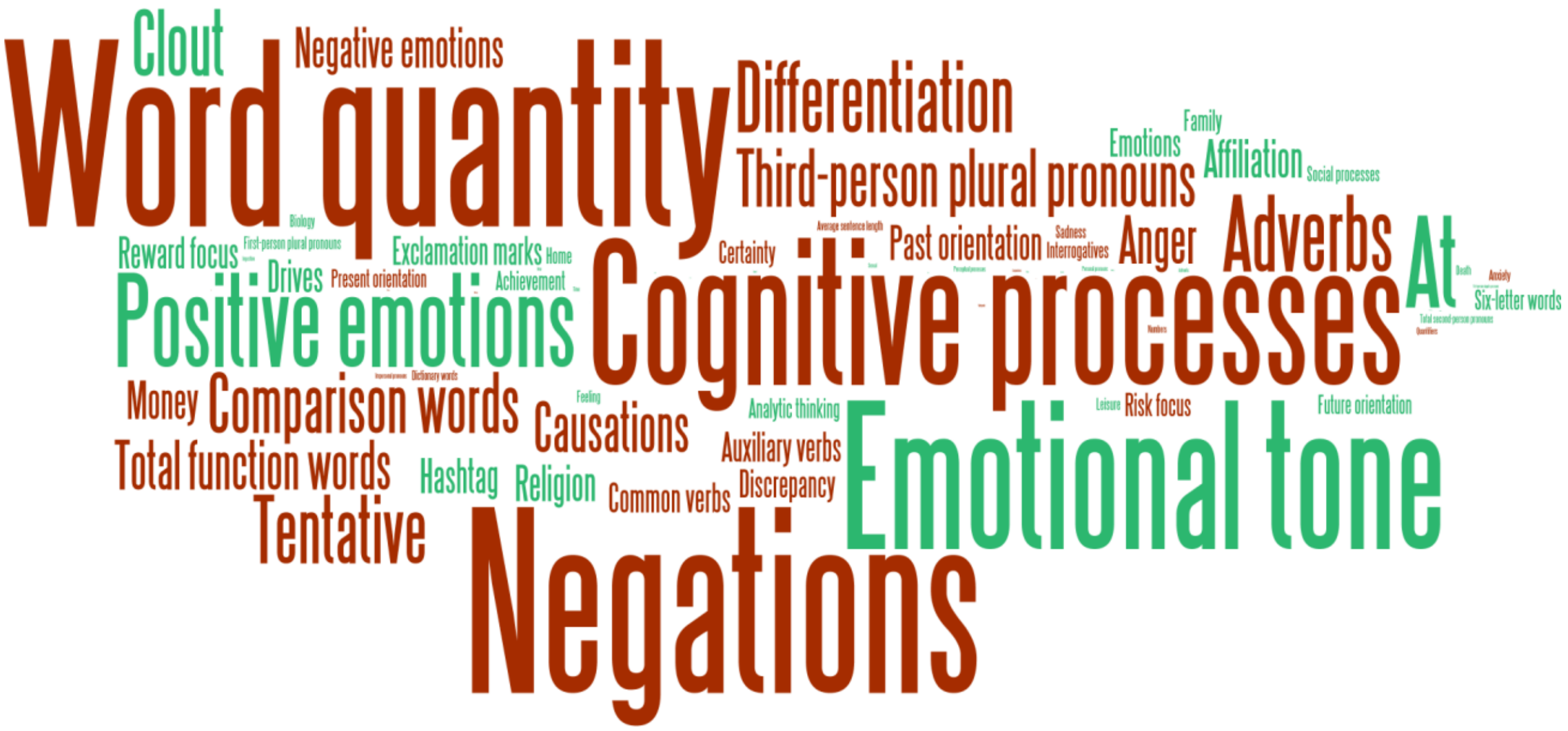

**Figure 1. Differences in variables between correct and incorrect tweets as a "Variable cloud".** The size of a variable is a function of the F statistics obtained in a MANOVA. Red/dark-colored variables had higher means for incorrect tweets, green/light-colored variables for correct tweets.

Next, we sought to find out how accurately we could identify the veracity of individual tweets based on these language cues. We estimated a logit model, starting with the 28 significant variables from the MANOVA. We removed variables stepwise to obtain the best value for the Akaike Information Criterion (AIC), i.e. until removing an additional parameter would decrease the log-likelihood of the model by more than one. We tried alternative approaches (see Methods section) but the approach reported here gave the best compromise between parsimony and goodness of fit. The final model comprises 10 variables. Table 1 displays the results in terms of marginal effects at the mean. For instance, an additional percentage point (pp) of money words increases the chance of an incorrect statement by 1.4pp whereas a 1pp increase in religious words decreases the chance of an incorrect statement by 6.5pp.

| | Marginal effects | Standard errors | z score | p-value |
|---|---|---|---|---|
| Word quantity | .005 | .001 | 3.948 | 0.000 |
| Emotional tone | -.002 | .001 | -3.909 | 0.000 |
| Third-person plural pronouns | .014 | .010 | 1.460 | 0.144 |
| Adverbs | .009 | .005 | 1.879 | 0.060 |
| Negation | .023 | .009 | 2.484 | 0.013 |
| Comparison words | .018 | .006 | 2.777 | 0.005 |
| Tentative | .012 | .009 | 1.425 | 0.154 |
| Money | .014 | .006 | 2.386 | 0.017 |
| Religion | -.065 | .033 | -1.961 | 0.050 |
| @ | -.168 | .047 | -3.547 | 0.000 |

**Table 1. Marginal effects of types of words and symbols on the probability of a tweet being incorrect.** Variables are expressed as percentage of total words, except word quantity and @ which are expressed as numbers of words / symbols. Logit regression, log-likelihood = -203.6, N=447.

The model summarized in Table 1 can predict the probabilities of each tweet being incorrect. We classified tweets as predicted incorrect if the model assigned a higher chance than we would *a priori* expect, i.e., higher than 29.53%, and as predicted correct otherwise. A hit occurs when an [in]correct statement is classified as predicted [in]correct. On Dataset 1, we obtained a hit rate of 76.51% for incorrect statements and 72.06% for correct statements, with an overall accuracy of 73.38%. Figure 2 displays the ROC curve: the hit rate for incorrect statements as a function of hit rates for correct statements for all possible cut-offs. The diagonal represents random guessing and the top-left corner perfect classification. For all cut-offs, our model clearly outperformed random guessing.

So far, we have demonstrated that there are language differences between the US president's factually correct and incorrect tweets. We then demonstrated that those language differences were strong enough to classify correct and incorrect tweets with an overall accuracy of 73.38%. Next, we investigate how consistent these differences in language use are by predicting the veracity of tweets in an out-of-sample test. We gathered a second dataset (Dataset 2) of 606 presidential tweets covering the three-month period November 2017 – January 2018 and applied the same screening methods, resulting in a dataset containing 464 tweets (22.84% incorrect). We then applied the model trained on Dataset 1 to identify how well we could classify tweets in Dataset 2. Using the same prior as for Dataset 1, we obtained a hit rate of 67.92% for incorrect statements and 74.02% for correct statements, with an overall accuracy of 72.63%. For comparison, random guessing using the prior probability of 29.53% for any tweet to be incorrect, would obtain an overall accuracy of 58.38% on Dataset 1 and of 61.11% on Dataset 2. The ROC curve for Dataset 2 is depicted in grey in Figure 2. It is lower than the curve for Dataset 1 but still clearly dominates random guessing. Alternatively, a cut-off of 50% would give an accuracy of 50% for random guessing and a higher overall accuracy for our model (77.40% for Dataset 1, 78.45% for Dataset 2). Hence, our approach using a prior probability as cut-off reduces the difference between random guessing and our model, representing a tougher test of our model.

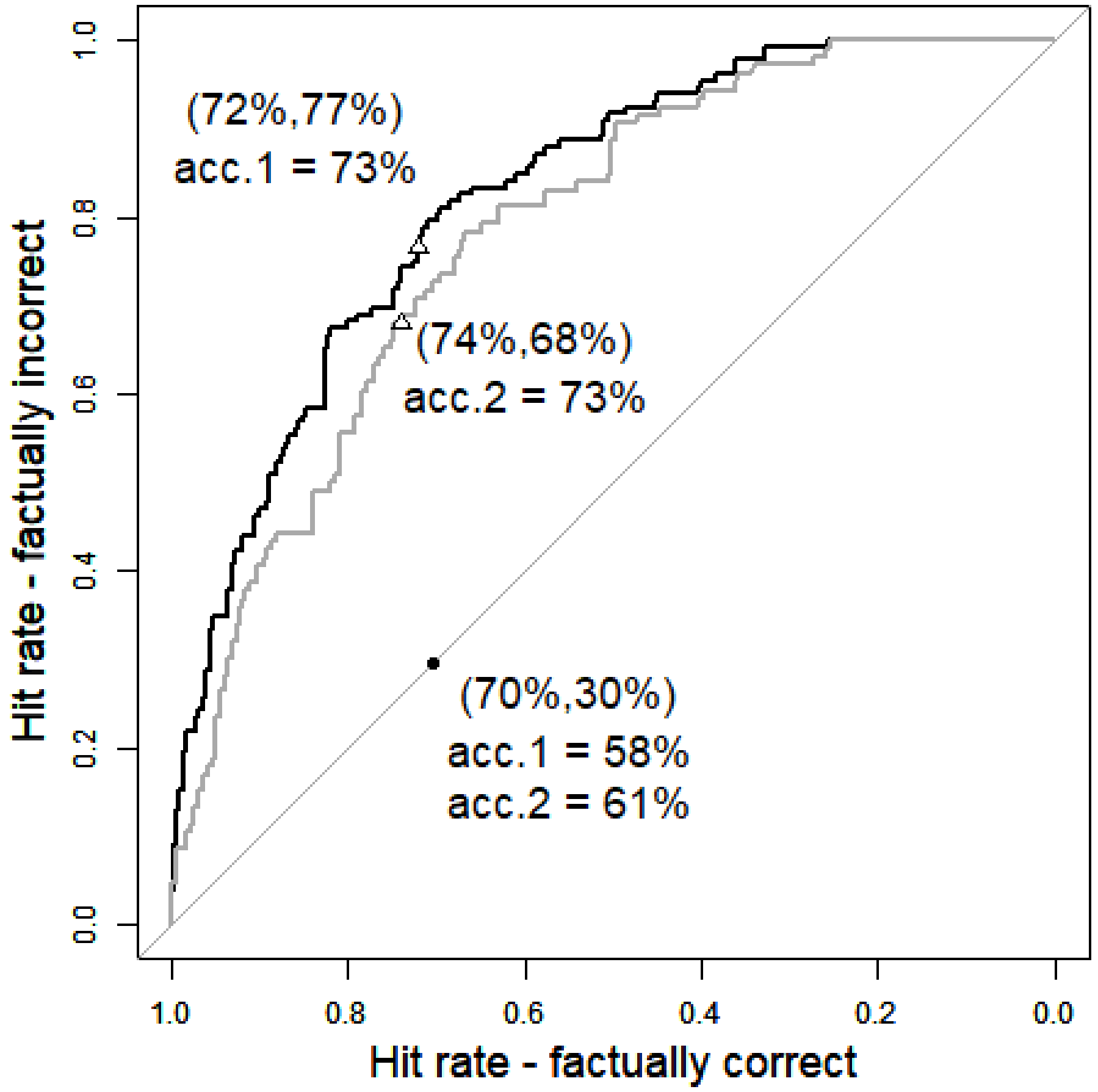

**Figure 2. Hit rates for factually correct statements as a function of hit rates for factually incorrect statement when cutoff varies.** Predictions are obtained from the model in Table 1. The black curve is for Dataset 1 and the grey for Dataset 2. The black dot represents the prior used as a random guess and triangles when the prior is used as cutoff in our model. Coordinates of the points, i.e. hit rates, are between brackets. Corresponding accuracies are denoted acc.1 for Dataset 1 and acc.2 for Dataset 2.

Previous research demonstrated that language changes when lying[5-7,13]. The majority of this research was conducted using single statements by large groups of people. Here, we investigated whether we could distinguish between correct and incorrect statements made by a single individual. Fact-checked tweets by the current US president uniquely allowed for such a comparison. The MANOVA results from this study showed large differences in language use between correct and incorrect tweets, supporting the deception hypothesis.

Some of these language differences are in line with previous findings in the deception literature. For example, liars experience more negative emotions (i.e., increased use of negative emotion and anger words, and negations), and they tend to distance themselves more (i.e., increased use of third person pronouns; *10*). The current US president further increased social distance by decreasing the use of @ and # symbols in incorrect tweets. However, his incorrect tweets were longer and more tentative than correct ones, whilst in general, lies tend to be shorter and contain fewer differentiating and tentative words than truths[13]. We cannot reject the explanation that longer tweets are simply more likely to contain erroneous statements than shorter tweets.

Compared to previous language-based deception studies, the 73%-74% overall accuracy of our model is promising. For example, Bond and colleagues[14] found classification rates of 63-67% based on the truthful and deceptive communications of several US presidential candidates. Our high out-of-sample accuracy shows that the language differences between true and false statements are relatively stable. Improving random guessing with 12-15pp when relying on priors, and 27-28pp when assuming an equal split between truths and lies, this model could help as a screening tool for anyone with an interest in fact-checking. Accuracy rates could potentially be further increased when combining multiple detection methods, such as keystroke analysis[24].

A limitation of the current study is that we used the fact-checking output of an independent third party as a proxy for the ground truth, potentially introducing noise in our data. For instance, the Washington Post only labelled statements as incorrect when they contained incorrect verifiable facts. Opinions and statements about the future are not labelled as factually incorrect, resulting in a conservative estimation of all incorrect tweets. While the ground truth cannot be guaranteed, focusing on statements that are factually verifiable made the classification less sensitive to the political opinion of the fact-checkers.

In this paper we demonstrated that the language people use can unintentionally reveal many things about them. Linguistic analysis can serve to test whether incorrect statements are predictable, and therefore likely to be deceptive. On the one side, such models can be a screening tool to help journalists in their work as the fourth pillar of democracy. They could also be used to identify potentially incorrect information on social media. On the other side, any person having access to people's personal posts and a way to approximate ground truth, could develop a model such as the one in this paper. Therefore, these results also constitute a warning to all posting a wealth of private information online.

## References and Notes

1. Ahmadian, S., Azarshahi, S., Paulhus, D. L. Explaining Donald Trump via communication style: Grandiosity, informality, and dynamism. *Pers. Indiv. Differ.* **107**, 49-53 (2017).
2. Lanning, K., Pauletti, R.E., King, L.A., McAdams, D. P. Personality development through natural language. *Nat Hum Behav*. **2**, 327-334 (2018).
3. Mehl, M. R. Gosling, S. D., Pennebaker, J. W. Personality in its natural habitat: Manifestations and implicit folk. Theories of personality in daily life. *J Pers Soc Pyschol*, **90**, 862-877 (2006).
4. Weintraub, W. *Verbal Behavior in Everyday Life* (New York, US, Springer, 1989).
5. Fuller, C. M., Twitchell, D. P., Biros, D. P., Wilson, R. L. Real-world deception and the impact on severity. *J Comput Inform Syst*. **55**, 59-67 (2015).
6. Bond, G. D., Lee, A. Y. Language of lies in prison: Linguistic classification of prisoners' truthful and deceptive natural language. *Appl. Cognitive Psych.* **19**, 313-329 (2005).
7. Newman, M. L., Pennebaker, J. W., Berry, D. S., Richards, J. M. Lying words: Predicting deception from linguistic styles. *Pers. Soc. Psychol. B.* **29**, 665–675 (2003).
8. DePaulo, B. M. et al. Cues to deception. *Psychol. Bull.* **129**, 74-118 (2003).
9. Pennebaker, J. W., Boyd, R. L., Jordan, K., Blackburn, K. *The Development and Psychometric Properties of LIWC2015* (University of Texas, Austin, US, 2015).
10. Salton, G. Automatic text analysis. *Science* **168,** 335-343 (1970).
11. Richardson, B. H., Taylor, P. J., Snook, B., Conchie, S. M. Bennell, C. Language style matching and police interrogation outcomes. *Law Human Behav.* **38**, 357-366 (2014).
12. Hirschberg, J., Manning, C. D. Advances in natural language processing. *Science* **349,** 261-266 (2015).
13. Hauch, V., Blandon-Gitlin, I., Masip, J., Sporer, S. L. Are computers effective lie detectors? A meta-analysis of linguistic cues to deception. *Pers. Soc. Psychol. Rev.* **19**, 307-342 (2015).
14. Bond, G. D. et al. 'Lyin' Ted', 'Crooked Hillary', and 'Deceptive Donald': Language of lies in the 2016 US Presidential debates. *Appl. Cognitive Psych.* **31**, 668-677 (2017).
15. Researchers have studied transcribed police interviews, trial documents of convicted criminals, and televised pleas for missing relatives, but even in those cases it remains unknown, on a statement level, which claims are true or false.
16. Hasen, R. L. A constitutional right to lie in campaigns and elections. *Mont. L. Rev.*, **74**, 53-77 (2013).
17. New York Times, Pithy, Mean and Powerful: How Donald Trump Mastered Twitter for 2016, https://www.nytimes.com/2015/10/06/us/politics/donald-trump-twitter-use-campaign-2016.html (5 October 2015).
18. Court ruling Case 1:17-cv-05205-NRB determined that it is against the constitution for @realDonaldTrump to block Twitter followers.

19. Alterman, E. *When Presidents Lie: A History of Official Deception and its Consequences* (Penguin, New York, US, 2004).
20. McGranahan, C. An anthropology of lying: Trump and the political sociality of moral outrage. *Am. Ethnol.* **44**, 243-248 (2017).
21. New Yorker, It's True: Trump Is Lying More, and He's Doing It on Purpose, https://www.newyorker.com/news/letter-from-trumps-washington/trumps-escalating-war-on-the-truth-is-on-purpose (3 August 2018).
22. Bernstein, D. M., Loftus, E. F. How to tell if a particular memory Is true or false. *Perspect on Psychol Sci*. **4**, 370-374 (2009).
23. Zuckerman, M. DePaulo, B. M., Rosenthal, R. Verbal and nonverbal communication of deception in *Advances in Experimental Social Psychology*, L. Berkowitz, Ed. (Academic Press, New York, US, 1981), vol. 14, pp. 1-57.
24. Monara, M. et al., Covert lie detection using keyboard dynamics. *Sci Rep.* **8**, 1976 (2018).

**Acknowledgments:** The authors would like to thank The Washington Post Fact Checker team for providing their fact-checked dataset of Trumps communications. We would also like to thank Benjamin Tereick for providing methodological suggestions for this paper, and Prof. Dr. Jozien Bensing and Dr. Annelies Vredeveldt for providing feedback on the written paper. Ethical approval was granted by the Internal Review Board for Non-Experimental Research of the Erasmus Research Institute of Management, Erasmus University Rotterdam. Figure 1 was made using http://www.wordle.net/advanced. **Funding**: European Research Council Starting Grant 638408 Bayesian Markets; **Author contributions:** Sophie Van Der Zee: Conceptualization, data curation, formal analysis, resources, writing – original draft, writing - review & editing; Ronald Poppe: data curation, formal analysis, software, validation, visualization, writing – original draft, writing - review & editing; Alice Havrileck: data curation, writing - review & editing; Aurélien Baillon: funding acquisition, formal analysis, writing – original draft, writing - review & editing; **Competing interests:** Authors declare no competing interests. **Data availability statement:** Upon publication, the complete datafile including screened tweets, veracity verdicts, and associated LIWC scores, and the R code to replicate all statistical analyses and ROC curves will be made available without restrictions on Github.

## Methods

### ***Datasets***

We collected a dataset (Dataset 1) of presidential tweets by @realDonaldTrump from http://www.trumptwitterarchive.com, covering a recent three-month period (February – April 2018). In total, 605 tweets were gathered. The Washington Post provided us with a dataset comprising fact-checked communications from this time period that we matched to our Twitter data file. As a result, we could identify which tweets were deemed factually incorrect, and labelled the remainder as correct. Subsequently, this dataset was screened to reduce noise. We removed retweets (66) and tweets containing quotes of over 6 words (52) to eliminate contamination of our data with tweets from others. We also removed duplicate tweets (16) and tweets solely containing web links (2). In addition, we merged tweets that together conveyed one message (22), leaving a final dataset of 447 tweets. Of these 447 tweets, 132 tweets (29.53%) were classified as factually incorrect and 315 as factually correct (70.47%).

Subsequently, we used the software program Linguistic Inquiry and Word Count (LIWC 2015) to identify how often different word categories were used in each tweet. LIWC calculates a score for 93 different categories, ranging from linguistic dimensions such as personal pronouns and negations, to psychological processes such as positive and negative emotions and cognitive processing[2]. During the screening process, we removed web links from the tweets, thereby altering punctuation. As a result, we excluded the LIWC punctuation categories except exclamation marks, leaving 82 categories. In addition, two dummy variables were created to identify whether a tweet included an @ and/or a #, since these symbols are specifically relevant to tweets.

We gathered a second dataset (Dataset 2) of presidential tweets covering the three-month period (November 2017 – January 2018) preceding the period covered in Dataset 1. In total, 606 tweets were gathered. We applied identical screening methods, removing retweets (85), duplications (6), tweets containing severe spelling mistakes (2), and tweets containing quotes with more than 6 words (29). In addition, 20 tweets were merged, leaving a final dataset of 464 tweets. Of these 464 tweets, 106 (22.84%) were classified as factually incorrect, leaving 358 factually correct tweets (77.16%).

## ***MANOVA***

We first ran a MANOVA to compare the means of the 82 LIWC variables and the two dummy variables (presence of @ or #) between the factually correct and the factually incorrect tweets. The Supplementary Information file "*SI File MANOVA.csv*" reports the results of the MANOVA on Dataset 1. It provides the means for factually correct and incorrect statements in Dataset 1, the F-statistics, p-values, and significance levels (*: $p < 0.05$; **: $p < 0.01$; ***: $p < 0.001$). In total, 38 variables were significant at a 5% level, among which 28 were significant at a 1% level, and among which 15 were significant at a 0.01% level.

## ***Model selection***

The second step of our analysis involved developing a model to predict factually incorrect statements. We used logit regressions with veracity as dependent variables and we had to determine which subset of the 84 variables to use. Too many variables may lead to overfitting and poor prediction power. To select the model (i.e., the set of explanatory variables), we use two criteria:

- The Akaike Information Criterion (AIC) which penalizes the log-likelihood (measuring the goodness of fit) by the number of variables to favor parsimonious models;
- The Area Under the Curve (AUC) of Receiver Operating Characteristic (ROC) curves. The logit model provides probabilistic predictions for each tweet. We can use these probabilities to classify tweets as "predicted correct" vs. "predicted incorrect". For instance, with a cut-off of 50%, we require that a tweet has a predicted probability of more than 50% to be classified as "predicted incorrect". A "hit" occurs when an [in]correct statement is classified as predicted [in]correct. ROC curves display hit rates for incorrect statements as a function of hit rates for correct statements when the cut-off varies. In other words, calling factually incorrect statements “positives”, ROC curves display the rate of true positives (also known as sensitivity or recall) as a function of 1 minus the rate of false positives (also known as specificity, the hit rates for negatives). By tradition, the x-axis is going from 1 to 0 if it displays specificity, and from 0 to 1 if it displays the rate of false positives. The diagonal represents random guessing for all possible cut-off values. The position of the curve above the diagonal shows the improvement over random guessing, with the upper left corner being perfect classification. The AUC measures the area under the curve. By construction, random guessing has an AUC of 0.5 and a perfect classifier of 1.

Several approaches were possible to select variables. When evaluating these approaches, we only considered the 28 variables that were significant at a 1% level according to the MANOVA because regressions of 447 observations on 84 variables led to perfect separation issues with extreme but non-significant coefficients.

First, we used stepwise selection of variables, either forward or backward, using AIC as a criterion to stop the process.

- Model “Forward”: The forward-stepwise selection means introducing variables one by one until the AIC does not decrease anymore. We implemented this approach starting with the variable with the highest F statistics in the MANOVA: Word quantity.
- Model “Backward”: The backward-stepwise approach starts with the whole set of 28 variables and removes them one by one, until the AIC does not decrease anymore.

Second, we also followed the approaches of[5-7, 14] and used the variables that were significant in the MANOVA. This gave us two possible models:

- Model “28 variables”, including the variables that were significant at 1% in the MANOVA.
- Model “15 variables” including the variables that were significant at 0.1% in the MANOVA.

Third, Least Absolute Shrinkage and Selection Operator (LASSO) is used in machine learning to select variables. It directly penalizes the presence of variables in the regression. We applied this approach as well:

- Model “Lasso”: variables selected by LASSO with the penalty level (known as lambda) that gives the lowest cross-validation mean standard errors.

We evaluated the models on Dataset 1 and obtained the following results:

- Model “Forward” (10 variables): Log-likelihood = -203.60; AIC = 429.20; AUC = 0.823.
- Model “Backward” (10 variables): Log-likelihood = -203.60; AIC = 429.20; AUC = 0.823.
- Model “28 variables”: Log-likelihood = -197.89; AIC = 453.78; AUC = 0.834.
- Model “15 variables”: Log-likelihood = -206.96; AIC = 445.92; AUC = 0.819.
- Model “Lasso” (25 variables): Log-likelihood = -194.74; AIC = 441.48; AUC = 0.841.

The “Forward” and “Backward” selection procedures gave the same model, with 10 variables. As could be expected, the model with all 28 variables has the highest log-likelihood. In terms of AIC, models “Forward” and “Backwards” score much better (i.e., get much lower AIC) than the other models, which is not surprising because they were developed using AIC as selection criterion. More interestingly, in terms of AUC, all models perform very closely from each other, with values 0.82-0.83meaning that they are all closer to perfect classification than to random guessing. To conclude, the “Backward” (or equivalently “Forward”) model manages to obtain comparable results than the other models in terms of AUC but with much more parsimony, using only 10 variables. This is the model reported in the paper and the remainder of the Supplementary Information file.

Out-of-Sample Predictions

Using the selected “Backwards” model, we could compute predicted probabilities on out-of-sample tweets. We used the coefficients obtained on the training set (Dataset 1) and applied them to the variables of the test set (Dataset 2) to predict the probability of each tweet to be factually incorrect. Using these predicted probabilities, we could again draw a ROC curve. Figure 2 in the paper displays the ROC curve obtained for the test set next to the ROC curve obtained for the training set.

The AUC for the test set is 0.782, to be compared with the AUC of 0.823 obtained for the training set. As could be expected, it decreases but is still very close to 0.8. To make the results more concrete, one may want to choose a specific cut-off, i.e. to decide which point of the curve to consider. Several approaches are possible.

- The first one is simply to use 0.5 but it has the drawback that it relies heavily on the exact quantitative prediction of our model, and for instance, on the logit assumption. It gives an accuracy of 77.40% [78.45%] for the training [test] set, and hit rates of 50.76% [38.68%] and 88.57% [90.22%] for factually incorrect and correct tweets, respectively. This threshold is very conservative, missing many incorrect statements.
- An alternative is to rely on the direction of predictions with respect to our prior. In other words, is a tweet more likely to be factually incorrect than we would have expected *ex ante*? To do so, used the base rate of factually incorrect statements in the training set (29.53%) and then used this cut-off to classify as “predicted incorrect” statements for which the model gave a higher probability than the base rate. We used this cut-off in the paper for its intuitive (Bayesian) appeal and its ease of interpretation.
- Other approaches would involve an external criterion (e.g., accuracy, average hit rates, average F1 score) and finding the cut-off that maximizes this criterion. There is no strict dominance of any criterion over all other possible criteria and such a choice would depend on practical or contextual arguments.

***Code availability***

Upon publication, the complete datafile including screened tweets, veracity verdicts, and associated LIWC scores, and the R code to replicate all statistical analyses and ROC curves will be made available without restrictions on Github.

**Supplementary Information**

| Variable name | LIWC name | Mean if correct | Mean if incorrect | F stat. | p-value | Sig. |
|---|---|---|---|---|---|---|
| Word quantity | Word_count | 31.61 | 44.35 | 45.81 | 0.0000 | * * * |
| Analytic thinking | Analytic | 74.67 | 68.65 | 4.88 | 0.0277 | * |
| Clout | Clout | 71.03 | 61.09 | 16.42 | 0.0001 | * * * |
| Authentic | Authentic | 32.91 | 29.71 | 0.94 | 0.3317 | |
| Emotional tone | Tone | 68.64 | 44.46 | 36.38 | 0.0000 | * * * |
| Average sentence length | WPS | 12.85 | 13.95 | 2.37 | 0.1243 | |
| Six-letter words | Sixltr | 22.79 | 20.45 | 4.97 | 0.0263 | * |
| Dictionary words | Dic | 79.24 | 81.05 | 2.42 | 0.1203 | |
| Total function words | function. | 43.34 | 47.34 | 12.21 | 0.0005 | * * * |
| Total pronouns | pronoun | 8.98 | 8.72 | 0.15 | 0.6986 | |
| Personal pronouns | ppron | 6.14 | 5.52 | 1.22 | 0.2694 | |
| First-person singular pronouns | i | 1.24 | 1.04 | 0.75 | 0.3877 | |
| First-person plural pronouns | we | 2.37 | 1.68 | 3.10 | 0.0788 | |
| Total second-person pronouns | you | 1.11 | 0.64 | 2.16 | 0.1423 | |
| Third-person singular pronouns | shehe | 0.69 | 0.68 | 0.00 | 0.9944 | |
| Third-person plural pronouns | they | 0.73 | 1.47 | 14.32 | 0.0002 | * * * |
| Impersonal pronouns | ipron | 2.84 | 3.21 | 1.32 | 0.2504 | |
| Articles | article | 6.42 | 6.67 | 0.32 | 0.5698 | |
| Prepositions | prep | 12.7 | 12.74 | 0.00 | 0.9442 | |
| Auxiliary verbs | auxverb | 7.58 | 9.09 | 7.34 | 0.0070 | * * |
| Adverbs | adverb | 3.66 | 5.49 | 18.86 | 0.0000 | * * * |
| Conjunctions | conj | 4.71 | 5.02 | 0.61 | 0.4368 | |
| Negations | negate | 1.04 | 2.48 | 45.55 | 0.0000 | * * * |
| Common verbs | verb | 13.46 | 15.5 | 7.38 | 0.0068 | * * |
| Adjectives | adj | 6.47 | 6.19 | 0.18 | 0.6718 | |
| Comparison words | compare | 1.37 | 2.49 | 14.98 | 0.0001 | * * * |
| Interrogatives | interrog | 0.91 | 1.32 | 3.98 | 0.0467 | * |
| Numbers | number | 1.62 | 2.13 | 1.86 | 0.1739 | |

**Table S.1. Statistics for the 28 LIWC categories**, with means for factual correct and incorrect statements in Dataset 1. F-statistics and significance values are reported. *: $p < 0.05$; **: $p < 0.01$; ***: $p < 0.001$.

| Variable name | LIWC name | Mean if correct | Mean if incorrect | F stat. | p-value | Sig. |
|---|---|---|---|---|---|---|
| Quantifiers | quant | 1.92 | 2.34 | 1.71 | 0.1911 | |
| Emotions | affect | 9.83 | 7.8 | 6.97 | 0.0086 | * * |
| Positive emotions | posemo | 7.56 | 4.06 | 23.10 | 0.0000 | * * * |
| Negative emotions | negemo | 2.19 | 3.64 | 9.99 | 0.0017 | * * |
| Anxiety | anx | 0.13 | 0.24 | 2.59 | 0.1084 | |
| Anger | anger | 0.35 | 0.79 | 11.91 | 0.0006 | * * * |
| Sadness | sad | 0.5 | 0.84 | 3.21 | 0.0740 | |
| Social processes | social | 9.52 | 8.15 | 3.82 | 0.0514 | |
| Family | family | 0.24 | 0.05 | 5.28 | 0.0220 | * |
| Friends | friend | 0.2 | 0.16 | 0.20 | 0.6513 | |
| Female references | female | 0.48 | 0.47 | 0.00 | 0.9693 | |
| Male references | male | 0.84 | 0.74 | 0.19 | 0.6633 | |
| Cognitive processes | cogproc | 6.96 | 10.53 | 36.05 | 0.0000 | * * * |
| Insight | insight | 0.99 | 1.05 | 0.10 | 0.7478 | |
| Causations | cause | 1.05 | 1.82 | 12.55 | 0.0004 | * * * |
| Discrepancy | discrep | 1.11 | 1.71 | 6.94 | 0.0087 | * * |
| Tentative | tentat | 1.16 | 2.09 | 16.87 | 0.0000 | * * * |
| Certainty | certain | 1.83 | 2.57 | 5.65 | 0.0179 | * |
| Differentiation | differ | 1.55 | 2.62 | 17.62 | 0.0000 | * * * |
| Perceptual processes | percept | 1.62 | 1.29 | 1.31 | 0.2525 | |
| Seeing | see | 0.88 | 0.72 | 0.43 | 0.5131 | |
| Hearing | hear | 0.33 | 0.4 | 0.39 | 0.5301 | |
| Feeling | feel | 0.4 | 0.17 | 2.91 | 0.0890 | |
| Biology | bio | 0.84 | 0.51 | 3.04 | 0.0819 | |
| Body | body | 0.13 | 0.06 | 0.79 | 0.3740 | |
| Health | health | 0.36 | 0.36 | 0.00 | 0.9610 | |
| Sexual | sexual | 0.03 | 0 | 1.12 | 0.2901 | |
| Ingestion | ingest | 0.15 | 0.06 | 1.14 | 0.2861 | |

**Table S.2. Statistics for the 28 LIWC categories**, with means for factual correct and incorrect statements in Dataset 1. F-statistics and significance values are reported. *: $p < 0.05$; **: $p < 0.01$; ***: $p < 0.001$.

| Variable name | LIWC name | Mean if correct | Mean if incorrect | F stat. | p-value | Sig. |
|---|---|---|---|---|---|---|
| Drives | drives | 13.24 | 10.89 | 8.00 | 0.0049 | * * |
| Affiliation | affiliation | 3.78 | 2.31 | 9.45 | 0.0022 | * * |
| Achievement | achieve | 2.25 | 1.58 | 4.76 | 0.0296 | * |
| Power | power | 5.02 | 5.19 | 0.12 | 0.7263 | |
| Reward focus | reward | 2.96 | 1.77 | 7.89 | 0.0052 | * * |
| Risk focus | risk | 0.82 | 1.35 | 5.73 | 0.0171 | * |
| Past orientation | focuspast | 2.52 | 3.58 | 8.59 | 0.0036 | * * |
| Present orientation | focuspresent | 8.65 | 10.03 | 4.45 | 0.0355 | * |
| Future orientation | focusfuture | 1.79 | 1.22 | 4.72 | 0.0303 | * |
| Relativity | relativ | 14.47 | 13.95 | 0.34 | 0.5588 | |
| Motion | motion | 1.61 | 1.57 | 0.03 | 0.8648 | |
| Space | space | 8.27 | 8.06 | 0.10 | 0.7542 | |
| Time | time | 5.09 | 4.51 | 1.20 | 0.2739 | |
| Job/Work | work | 5.08 | 5.15 | 0.01 | 0.9178 | |
| Leisure | leisure | 0.64 | 0.31 | 2.97 | 0.0854 | |
| Home | home | 0.46 | 0.23 | 4.04 | 0.0450 | * |
| Money | money | 1.25 | 2.26 | 9.42 | 0.0023 | * * |
| Religion | relig | 0.69 | 0.05 | 9.00 | 0.0029 | * * |
| Death | death | 0.39 | 0.16 | 2.31 | 0.1290 | |
| Informal | informal | 0.28 | 0.22 | 0.17 | 0.6843 | |
| Swear words | swear | 0.01 | 0.02 | 0.01 | 0.9207 | |
| Netspeak | netspeak | 0.07 | 0.11 | 0.63 | 0.4273 | |
| Assent | assent | 0.1 | 0.03 | 0.49 | 0.4862 | |
| Nonfluencies | nonflu | 0.08 | 0.08 | 0.03 | 0.8721 | |
| Fillers | filler | 0.01 | 0 | 0.42 | 0.5180 | |
| Exclamation marks | Exclam | 5.24 | 2.98 | 7.08 | 0.0081 | * * |
| # | | 0.12 | 0.03 | 8.61 | 0.0035 | * * |
| @ | | 0.22 | 0.04 | 22.56 | 0.0000 | * * * |

**Table S.3. Statistics for the 26 LIWC categories and two symbols (@ and #)**, with means for factual correct and incorrect statements in Dataset 1. F-statistics and significance values are reported. *: $p < 0.05$; **: $p < 0.01$; ***: $p < 0.001$.